\documentclass[runningheads]{llncs}
\usepackage{graphicx}
\usepackage{mathrsfs}
\usepackage{subfigure}
\usepackage{balance}

\usepackage{array}
\usepackage{booktabs}
\usepackage{microtype}
\usepackage{url}
\usepackage{amsmath}
\usepackage{siunitx}
\usepackage{color}
\usepackage{multirow}
\usepackage{booktabs}
\usepackage{float}
\usepackage{bbding}
\usepackage{hyperref}

\usepackage{times}
\usepackage{latexsym}
\usepackage{amsmath}
\usepackage{amssymb}
\usepackage{amsfonts} % For \mathbb command
\usepackage{bm} % For bold math symbols if needed (optional)

\newcommand{\vect}[1]{\mathbf{#1}}

\usepackage{bm}
\usepackage{booktabs}
\usepackage{amssymb}% http://ctan.org/pkg/amssymb
\usepackage{colortbl}
\usepackage{multirow}
\usepackage{color}
\usepackage{arydshln}
\usepackage[utf8]{inputenc}

\usepackage{microtype}

\usepackage{inconsolata}

\definecolor{DarkGreen}{RGB}{0, 128, 0}
\definecolor{deltap}{RGB}{119, 139, 204}
\definecolor{deltan}{RGB}{255, 129, 90}

\usepackage{xspace}
\usepackage{url}

\usepackage[misc]{ifsym}

\usepackage{tabularx}
\usepackage{booktabs}
\usepackage{makecell}
\usepackage{multirow}
\usepackage{boldline}
\usepackage{caption}

\usepackage{wrapfig}
\usepackage{atbegshi}

\usepackage{algorithm}
\usepackage{algorithmic}

\usepackage{xcolor}

\begin{document}
\newcommand{\modelname}{{CIRN}}

\title{Using External knowledge to Enhanced PLM for Semantic Matching}

\author{Min Li \and
Chun Yuan
}
\authorrunning{Li et al.}
\institute{Tsinghua Shenzhen International Graduate School, Tsinghua University, Beijing, China}

%\institute{Princeton University, Princeton NJ 08544, USA \and
%Springer Heidelberg, Tiergartenstr. 17, 69121 Heidelberg, Germany
\maketitle              % typeset the header of the contribution

\begin{abstract}
Modeling semantic relevance has always been a challenging and critical task in natural language processing. In recent years, with the emergence of massive amounts of annotated data, it has become feasible to train complex models, such as neural network-based reasoning models. These models have shown excellent performance in practical applications and have achieved the current state-of-the-art performance.
However, even with such large-scale annotated data, we still need to think: Can machines learn all the knowledge necessary to perform semantic relevance detection tasks based on this data alone? If not, how can neural network-based models incorporate external knowledge into themselves, and how can relevance detection models be constructed to make full use of external knowledge?
In this paper, we use external knowledge to enhance the pre-trained semantic relevance discrimination model. Experimental results on 10 public datasets show that our method achieves consistent improvements in performance compared to the baseline model.

\end{abstract}

\section{Introduction}
\label{sec:intro}
Semantic Relevance Modeling (SRM) stands as a cornerstone in Natural Language Processing (NLP). Its core objective is to assess the semantic relationship between two sentences. In tasks like paraphrase identification, SRM determines if two sentences convey the same meaning; in natural language inference, it evaluates whether a hypothesis can be deduced from a premise; and in answer sentence selection, it gauges the relevance of query-answer pairs to rank potential answers.

Historically, semantic matching research has predominantly followed two paths. One approach encodes sentences into low-dimensional vectors in a latent space and then applies a function to compute matching scores \cite{conneau2017supervised}. The other focuses on aligning phrases, aggregating information to predict similarity at the sentence level \cite{chen2016enhanced,tay2017compare}.
The advent of large-scale pre-trained language models (PLMs) has spurred efforts to incorporate external knowledge \cite{miller1995wordnet}. For instance, SemBERT \cite{zhang2020semantics} enhances BERT by concatenating semantic role annotations, UERBERT \cite{xia2021using} injects synonym knowledge, and SyntaxBERT \cite{bai2021syntax} integrates syntax trees. These practices have proven beneficial for various NLP tasks \cite{bowman2016fast}.

Despite neural networks' success in semantic relevance modeling with abundant training data, they often assume all necessary inference knowledge can be learned from the data during end-to-end training. In this study, we challenge this assumption and explore the role of external knowledge in semantic modeling. Consider the following example: when determining the relationship between two sentences where the semantic link between words like "wheat" and "corn" is absent from the training data or pre-training, models struggle to identify semantic contradictions. We posit that complex NLP tasks such as SRM can benefit from human-accumulated knowledge, especially when models cannot learn certain information independently.
We argue that external knowledge is crucial for SRM. Existing PLM training methods mainly optimize co-occurrence probabilities, while external structured knowledge can augment PLMs by enhancing feature interactions. Leveraging an external knowledge base like WordNet \cite{miller1995wordnet} to create a prior matrix and calibrate attention weights can improve PLM performance. This approach raises two key questions:

\textbf{Q1: How can we construct a prior matrix incorporating external knowledge?} We explore integrating knowledge about synonyms, antonyms, hypernyms, and hyponyms into PLMs. Such knowledge can aid in soft-aligning sentence pairs, capturing fine-grained differences, and modeling semantic contradictions.

\textbf{Q2: How should we integrate the signals from the prior matrix?} To fully exploit external knowledge, we use the prior matrix to calibrate attention alignment. However, simple fusion of prior and semantic signals may compromise PLMs' representational capabilities. To address this, we propose an adaptive fusion module. It first aligns the two signals using different attention mechanisms, then selectively extracts meaningful information via multiple gates, and finally scales the output to mitigate potential noise, enabling effective integration of semantic and external knowledge signals.

Our contributions are threefold. First, we detail methods for constructing and injecting prior matrices into PLMs. Second, we introduce a novel external knowledge calibration and fusion network that combines semantic and external knowledge, enhancing model interpretability. Finally, extensive experiments on 10 public datasets demonstrate that our approach outperforms strong baselines, validating its effectiveness.

\section{Related Work}
\label{sec:related_works}

Semantic relevance modeling has been a central research area in NLP, evolving significantly over time. The approaches can be broadly categorized into traditional neural network-based methods and pre-trained language model (PLM)-based methods.

\subsection{Neural Language Inference Approaches}
\subsubsection{Traditional Neural Network-based Methods}
Early semantic relevance modeling (SRM) relied on techniques such as syntactic feature extraction, text transformations, and relation analysis \cite{zheng2022robust}. While effective for specific tasks, these methods suffered from limited generalization. 
The emergence of large-scale annotated datasets \cite{bowman2015large} and deep learning algorithms propelled the development of neural network models for SRM. Attention mechanisms brought a major breakthrough, enabling models to capture alignment and dependency between sentences, thus uncovering semantic similarities beyond surface-level lexical matches \cite{conneau2017supervised}. Sentence-encoding methods mapped sentences to vector representations, followed by classification to determine semantic relationships \cite{conneau2017supervised}. Joint methods, leveraging cross-features via attention, improved performance by focusing on word and phrase alignments \cite{ma2022searching}. Architectural enhancements, like residual connections, facilitated deeper network training while preserving lower-level information \cite{liu2016learning}. 
Researchers employed various neural network architectures for semantic representation. For example, \cite{liang2019adaptive} emphasized sequential and semantic interdependencies, while \cite{xu2020enhanced} used convolutional filters for local context extraction. Attention-based methods, such as those in \cite{liu2023time}, extracted key sentence components and aligned sentence pairs. \cite{liu2023local} applied stacked Bi-LSTM with alignment factorization for multi-level feature analysis. Convolutional Neural Networks (CNNs) focused on local context, Recurrent Neural Networks (RNNs) captured sequential information, and models like \cite{fei2022cqg} used CNN-based multi-layer encoding for efficient phrase matching. \cite{dong2020distilling} exploited graph neural networks (GNNs) to model sentence structures, and \cite{xue2023dual} encoded sentences independently for similarity calculation.

\subsubsection{Pre-trained Language Model-based Methods}
The advent of PLMs, especially BERT \cite{devlin2018bert}, transformed SRM by offering powerful sentence representations through self-supervised learning on extensive corpora. This enabled effective transfer learning across diverse NLP tasks.
To boost PLM performance, researchers modified input encodings and pre-training tasks. XLNet \cite{wu2025unleashing} reduced the gap between pre-training and downstream tasks, and models like RoBERTa \cite{wu2024tablebench} and CharBERT \cite{ma2020charbert,wang-etal-2022-dabert} introduced notable advancements. Some works enhanced performance by using cross-features for alignment \cite{liang2019asynchronous}, while others, like DenseNet \cite{xue2024question}, leveraged recurrent and co-attentive information.
There is a growing trend of integrating explicit NLP knowledge into models. For instance, \cite{song-etal-2022-improving-semantic} enhanced sentence representations with syntactic dependencies. Knowledge-enhanced matching models adapted to interaction-based architectures: MIX \cite{li2024local} used POS and named-entity tags as prior features; SemBERT \cite{zhang2020semantics} augmented BERT with semantic role annotations; UERBERT \cite{gui2018transferring} injected synonym knowledge; and SyntaxBERT \cite{bai2021syntax} integrated syntax trees. However, existing models often struggle to capture fine-grained semantic differences in text pairs with high lexical similarity.

\subsection{Robustness Testing}
Despite neural network models' impressive performance in many tasks, they remain vulnerable in real-world applications \cite{gui2021textflint}. Minor textual changes can lead to incorrect predictions, especially when fine-grained semantic discrimination is required. Moreover, relying on a single evaluation metric often overestimates model capabilities and overlooks robustness details.
Recent research has focused on multi-faceted robustness evaluation. Tools like TextFlint \cite{gui2021textflint} conduct comprehensive analysis through various text transformations. \cite{liu2024resolving} established benchmarks for adversarial attacks, and \cite{li2024comateformer} proposed a more elaborate evaluation system with detailed metrics, advancing the understanding of model robustness.

\section{Knowledge-Infused Attention Mechanism}
\label{sec:model}

This section details our proposed model architecture, focusing on how external lexical knowledge is encoded and integrated into a PLM's attention mechanism to enhance Semantic Relevance Modeling (SRM).

\subsection{External Knowledge Representation}
\label{subsec:knowledge_representation}

We explore integrating explicit lexical-semantic relationships, often absent in standard PLM pre-training, to improve SRM. Knowledge about \textit{synonymy}, \textit{antonymy}, \textit{hypernymy}, and \textit{hyponymy} between words can aid alignment and reasoning. For instance, synonymy can facilitate soft alignment between sentences, antonymy can signal contradiction, and hypernym/hyponym relations can help capture semantic similarity or fine-grained differences.

We represent the external lexical knowledge between a word $w_i$ (from the first input text, e.g., premise) and a word $w_j$ (from the second input text, e.g., hypothesis) as a knowledge vector $\vect{k}_{ij} \in \mathbb{R}^{d_k}$. This vector could be a one-hot encoding of relation types derived from a knowledge base like WordNet \cite{miller1995wordnet}, or a learned embedding based on these relations.

\subsection{Knowledge-Aware Co-Attention}
\label{subsec:knowledge_co_attention}

To leverage this external knowledge during the initial interaction between the two input texts, we propose a knowledge-aware co-attention mechanism. Let the contextualized hidden states for the two input texts be $\vect{H}^A = [\vect{h}^A_1, ..., \vect{h}^A_m] \in \mathbb{R}^{d_h \times m}$ and $\vect{H}^B = [\vect{h}^B_1, ..., \vect{h}^B_n] \in \mathbb{R}^{d_h \times n}$, where $d_h$ is the hidden dimension, and $m, n$ are the sequence lengths. The initial similarity score $s_{ij}$ between $\vect{h}^A_i$ and $\vect{h}^B_j$ is augmented with the knowledge signal:
\begin{equation}
s_{ij} = (\vect{h}^A_i)^\mathrm{T} \vect{h}^B_j + G(\vect{k}_{ij})
\label{eq:similarity_score}
\end{equation}
where $G(\cdot)$ is a function mapping the knowledge vector to a scalar relevance score. We employ a simple yet effective mapping: $G(\vect{k}_{ij}) = \gamma \mathbb{I}(\vect{k}_{ij})$, where $\gamma$ is a tunable hyperparameter and $\mathbb{I}(\cdot)$ is an indicator function:
\begin{align}
\mathbb{I}(\vect{k}_{ij}) =
  \begin{cases}
    1  & \quad \text{if } \vect{k}_{ij} \text{ indicates any relevant relation;} \\
    0  & \quad \text{otherwise (e.g., } \vect{k}_{ij} = \vect{0} \text{ or no relation found).}
  \end{cases}
\end{align}
This formulation boosts the attention score for word pairs possessing a known semantic relationship, promoting their alignment. While more complex functions $G$ (e.g., an MLP) are possible, we found this simple indicator effective in practice.

Based on these augmented similarity scores $s_{ij}$, we compute standard attention-based context vectors. The attention weights $\omega^A_{ij}$ representing the relevance of $\vect{h}^B_j$ to $\vect{h}^A_i$ and the resulting context vector $\vect{c}^A_i$ are:
\begin{align}
\omega^A_{ij} & = \frac{\exp(s_{ij})}{\sum_{p=1}^{n}\exp(s_{ip})} \,, & \vect{c}^A_i =\sum_{j=1}^{n}\omega^A_{ij} \vect{h}^B_j
\label{eq:context_A}
\end{align}
Similarly, the weights $\omega^B_{ij}$ for the relevance of $\vect{h}^A_i$ to $\vect{h}^B_j$ and context vector $\vect{c}^B_j$ are:
\begin{align}
\omega^B_{ij} & = \frac{\exp(s_{ij})}{\sum_{p=1}^{m}\exp(s_{pj})} \,, & \vect{c}^B_j =\sum_{i=1}^{m}\omega^B_{ij} \vect{h}^A_i
\label{eq:context_B}
\end{align}
These context vectors capture the initial cross-text alignment informed by both semantics and external knowledge. We can define an aggregate prior knowledge matrix $\vect{K}_{prior} \in \mathbb{R}^{m \times n}$ based on these weights, for example, by averaging: $\vect{K}_{prior}[i, j] = (\omega^A_{ij} + \omega^B_{ij}) / 2$. (Note: The original text used $\mathbf{M_F}(i,j)$ based on $\alpha_{ij}$ and $\beta_{ij}$ which directly used $e_{ij}$ in the denominator - the formulation above defines a similar matrix based on the augmented scores $s_{ij}$). For integration into self-attention later, we assume $\vect{K}_{prior}$ is appropriately formatted, potentially focusing on self-attention within a concatenated sequence or adapted based on the specific PLM layer. Let's assume $\vect{K}_{prior} \in \mathbb{R}^{L \times L}$ where $L$ is the sequence length for the self-attention layer.

\subsection{Integrating Prior Knowledge into Self-Attention}
\label{subsec:integration_self_attention}

We inject the prior knowledge represented by $\vect{K}_{prior}$ directly into the PLM's self-attention mechanism. Standard scaled dot-product attention computes attention weights between queries ($\vect{Q}$) and keys ($\vect{K}$) and applies these weights to values ($\vect{V}$). We modify this to incorporate our knowledge prior. Within each attention head, we compute two parallel attention outputs:

\begin{equation}
\begin{aligned}
\text{Scores}_{sem} &= \frac{\vect{Q} \vect{K}^T}{\sqrt{d_k}} \\
\vect{O}_{sem} &= \text{softmax}(\text{Scores}_{sem}) \vect{V} \\
\text{Scores}_{knw} &= \frac{\vect{Q} \vect{K}^T \odot \vect{K}_{prior}}{\sqrt{d_k}} \\
\vect{O}_{knw} &= \text{softmax}(\text{Scores}_{knw}) \vect{V}
\end{aligned}
\label{eq:modified_attention}
\end{equation}
where $\vect{Q}, \vect{K} \in \mathbb{R}^{L \times d_k}$, $\vect{V} \in \mathbb{R}^{L \times d_v}$, $L$ is the sequence length, $d_k, d_v$ are key/value dimensions. $\odot$ denotes element-wise multiplication, assuming $\vect{K}_{prior}$ is broadcastable or reshaped appropriately to match the attention score dimensions ($L \times L$). $\vect{O}_{sem} \in \mathbb{R}^{L \times d_v}$ is the standard semantic attention output, while $\vect{O}_{knw} \in \mathbb{R}^{L \times d_v}$ is the output modulated by the external knowledge prior.

\subsection{Adaptive Fusion Module}
\label{subsec:adaptive_fusion}

Simply combining $\vect{O}_{sem}$ and $\vect{O}_{knw}$ (e.g., concatenation or summation) might not effectively leverage their distinct information and could introduce noise from imperfect prior knowledge. We propose an adaptive fusion module to dynamically integrate these two signals.

First, we facilitate mutual interaction between the semantic and knowledge-modulated signals. Let $\vect{o}^{sem}_i$ and $\vect{o}^{knw}_i$ be the $i$-th row vectors (representing the $i$-th token's output) from $\vect{O}_{sem}$ and $\vect{O}_{knw}$, respectively. We update the knowledge signal using semantic guidance, and vice-versa:

{\vspace{-0.3cm}
\setlength{\abovedisplayskip}{0.1cm}
\setlength{\belowdisplayskip}{0.2cm}
\begin{equation}
\begin{aligned}
% Updating knowledge signal using semantic guidance
\vect{z}^{knw}_i &= \tanh(\vect{W}_{O^{knw}} \vect{O}_{knw}^T \oplus (\vect{W}_{o^{sem}} \vect{o}^{sem}_i + \vect{b}_{o^{sem}})) \\
\alpha^{knw}_i &= \text{softmax}(\vect{w}_{z^{knw}}^T \vect{z}^{knw}_i + b_{z^{knw}}) \\ % Attention weights (scalar per position)
\hat{\vect{o}}^{knw}_i &= \sum_{j=1}^L \alpha^{knw}_{ij} \vect{o}^{knw}_j \\ % Weighted sum - Check dimensions carefully - original had Pri * softmax(...)
% Updating semantic signal using updated knowledge guidance
\vect{z}^{sem}_i &= \tanh(\vect{W}_{O^{sem}} \vect{O}_{sem}^T \oplus (\vect{W}_{\hat{o}^{knw}} \hat{\vect{o}}^{knw}_i + \vect{b}_{\hat{o}^{knw}})) \\
\alpha^{sem}_i &= \text{softmax}(\vect{w}_{z^{sem}}^T \vect{z}^{sem}_i + b_{z^{sem}}) \\
\hat{\vect{o}}^{sem}_i &= \sum_{j=1}^L \alpha^{sem}_{ij} \vect{o}^{sem}_j % Weighted sum - Check dimensions carefully
\end{aligned}
\label{eq:mutual_attention}
\end{equation}}
\textit{(Self-correction: The original equations using `underline` and matrix multiplication seemed overly complex/potentially incorrect for standard attention. The above uses a more typical cross-attention pattern where one signal attends over the other. Ensure weight dimensions match. $\vect{W}_{O^{knw}}, \vect{W}_{O^{sem}} \in \mathbb{R}^{d_{attn} \times d_v}$; $\vect{W}_{o^{sem}}, \vect{W}_{\hat{o}^{knw}} \in \mathbb{R}^{d_{attn} \times d_v}$; $\vect{w}_{z^{knw}}, \vect{w}_{z^{sem}} \in \mathbb{R}^{d_{attn}}$; $\oplus$ implies concatenation then linear layer, adjust dimensions accordingly. Resulting $\hat{\vect{o}}$ vectors are in $\mathbb{R}^{d_v}$. Adjust if the original intent was different.)}

Second, we adaptively fuse these refined features $\hat{\vect{o}}^{sem}_i$ and $\hat{\vect{o}}^{knw}_i$ using a gating mechanism:

{\vspace{-0.3cm}
\setlength{\abovedisplayskip}{0.1cm}
\setlength{\belowdisplayskip}{0.2cm}
\begin{equation}
\begin{aligned}
\tilde{\vect{o}}^{knw}_i &= \tanh(\vect{W}_{\tilde{o}^{knw}} \hat{\vect{o}}^{knw}_i + \vect{b}_{\tilde{o}^{knw}}) \\
\tilde{\vect{o}}^{sem}_i &= \tanh(\vect{W}_{\tilde{o}^{sem}} \hat{\vect{o}}^{sem}_i + \vect{b}_{\tilde{o}^{sem}}) \\
g^{fuse}_i &= \sigma(\vect{W}_{g^{fuse}}[\tilde{\vect{o}}^{knw}_i ; \tilde{\vect{o}}^{sem}_i] + b_{g^{fuse}}) \\
\vect{u}_i &= g^{fuse}_i \odot \tilde{\vect{o}}^{sem}_i + (1 - g^{fuse}_i) \odot \tilde{\vect{o}}^{knw}_i
\end{aligned}
\label{eq:gated_fusion}
\end{equation}}
where $\vect{W}_{\tilde{o}^{knw}}, \vect{W}_{\tilde{o}^{sem}} \in \mathbb{R}^{d_{hidden} \times d_v}$; $\vect{W}_{g^{fuse}} \in \mathbb{R}^{1 \times 2d_{hidden}}$. The gate $g^{fuse}_i$ adaptively balances the contribution of the refined semantic and knowledge signals to form a unified representation $\vect{u}_i \in \mathbb{R}^{d_{hidden}}$.

Third, acknowledging potential noise in the fused signal or the external knowledge itself, we introduce a final filtration gate. This gate selectively incorporates the fused information $\vect{u}_i$ based on its relevance, potentially modulated by the original semantic signal $\vect{o}^{sem}_i$:
{\vspace{-0.3cm}
\setlength{\abovedisplayskip}{0.1cm}
\setlength{\belowdisplayskip}{0.2cm}
\begin{equation}
\begin{aligned}
g^{filter}_i &= \sigma(\vect{W}_{g^{filter}}[\vect{o}^{sem}_i ; \vect{u}_i] + b_{g^{filter}}) \\
\vect{y}_i &= g^{filter}_i \odot \tanh(\vect{W}_{y} \vect{u}_i + \vect{b}_{y})
\end{aligned}
\label{eq:filtration_gate}
\end{equation}}
where $\vect{W}_{g^{filter}} \in \mathbb{R}^{1 \times (d_v + d_{hidden})}$; $\vect{W}_{y} \in \mathbb{R}^{d_v \times d_{hidden}}$. The final representation $\vect{y}_i \in \mathbb{R}^{d_v}$ adaptively integrates the knowledge-infused signal. This vector $\vect{y}_i$ (or the sequence $\vect{Y} = [\vect{y}_1, ..., \vect{y}_L]$) then replaces the standard output of the self-attention layer and is passed to subsequent layers (e.g., feed-forward network) in the Transformer block.

\begin{table*}
\centering
\caption{\label{citation-guide-gule}
The performance comparison of our model with other methods. We report Accuracy $\times$ 100 on 6 GLUE datasets. Methods with $\dagger$ indicate the results from their papers, while methods with $\ddagger$ indicate our implementation.
}
\renewcommand\arraystretch{1.2}
\scalebox{0.8}{
\setlength{\tabcolsep}{1.6mm}{
\begin{tabular}{lcccccccc}
\toprule
\midrule
\multirow{2}*{Model} &\multirow{2}*{Pre-train} & \multicolumn{3}{c}{Sentence Similarity} &\multicolumn{3}{c}{Sentence Inference} &\multirow{2}*{Avg}\\  \cmidrule(r){3-8} 
         ~& &\text{MRPC} & \text{QQP} & \text{SST-B} &\text{MNLI-m/mm}  & \text{QNLI} & \text{RTE}  \\

%\text{Method} & \text{MRPC} & \text{QQP} & \text{MNLI-m/mm}& \text{QNLI} & \text{RTE} & \text{SST-B} & \text{Avg} \\
\midrule
%\multicolumn{7}{c}{\textbf{No Pre-trained Models}} \\
% \textbf{BiLSSM$\dagger$\cite{hochreiter1997long}} & 76.2 & 79.2 & 67.0/67.6 & 76.5 & 52.3 & - & - \\
\text{BiMPM$\dagger$} & \XSolidBrush & 79.6 & 85.0 & - & 72.3/72.1 & 81.4 & 56.4  & - \\
\text{CAFE$\dagger$\cite{tay2017compare}}& \XSolidBrush  & 82.4 & 88.0 & - & 78.7/77.9 & 81.5 & 56.8  & - \\
\text{ESIM$\dagger$\cite{chen2016enhanced}}& \XSolidBrush & 80.3 & 88.2 & - & - & 80.5 & -  & - \\
\text{Transformer$\dagger$\cite{vaswani2017attention}}& \XSolidBrush  & 81.7 & 84.4 & 73.6 & 72.3/71.4 & 80.3 & 58.0  & 74.53 \\
\midrule
%\multicolumn{7}{c}{\textbf{Pre-trained Models}} \\
\text{BiLSTM+ELMo+Attn$\dagger$} &\Checkmark  & 84.6 & 86.7 & 73.3 & 76.4/76.1 & 79.8 & 56.8  & 76.24 \\
\text{OpenAI GPT$\dagger$} &\Checkmark  & 82.3 & 70.2 & 80.0 & 82.1/81.4 & 87.4 & 56.0  & 77.06 \\
\text{UERBERT$\ddagger$\cite{xia2021using}} &\Checkmark  & 88.3 & 90.5 & 85.1 & 84.2/83.5 & 90.6 & 67.1 & 84.19 \\
% \text{MT-DNN$\ddagger$\cite{liu2019multi}} & 88.5 & 90.6 & 84.6/84.4 & - & 70.0 & 88.3 & 84.4 \\
\text{SemBERT$\dagger$\cite{zhang2020semantics}} &\Checkmark   & 88.2 & 90.2 & 87.3 & 84.4/84.0 & 90.9 & 69.3  & 84.90 \\
\midrule
\text{BERT-base$\ddagger$\cite{devlin2018bert}}&\Checkmark  & 87.2 & 89.0 & 85.8 & 84.3/83.7 & 90.4 & 66.4  & 83.83 \\
\text{SyntaxBERT-base$\dagger$\cite{bai2021syntax}}&\Checkmark  & \textbf{89.2} & 89.6 & 88.1 & 84.9/84.6 & 91.1 & 68.9  & 85.20 \\
\textbf{Ours(Bert-base)}$\ddagger$&\Checkmark  & 89.0 & \textbf{91.2} & \textbf{88.2}  & \textbf{85.1}/\textbf{84.9} & \textbf{91.3} & \textbf{69.8} & \textbf{85.49} \\
\midrule
\text{BERT-large$\ddagger$\cite{devlin2018bert}}&\Checkmark  & 89.3 & 89.3 & 86.5 & 86.8/85.9 & 92.7 & 70.1  & 85.80 \\
\text{SyntaxBERT-large$\dagger$\cite{bai2021syntax}}&\Checkmark  & \textbf{92.0} & 89.5 & 88.5 & 86.7/86.6 & 92.8 & 74.7  & 87.26 \\
\textbf{Ours(Bert-large)}$\ddagger$ &\Checkmark & 91.2 & \textbf{91.8} & \textbf{89.6}& \textbf{87.2}/\textbf{86.9} & \textbf{94.6} & \textbf{75.2}  & \textbf{88.11} \\
\midrule
\bottomrule
\end{tabular}}}

\vspace{-0.2cm}
\end{table*}

\begin{table}[th]
\centering
\caption{\label{performance4}
The performance on 4 other datasets, including SNLI, Scitail(Sci), SICK and TwitterURL(Twi).}
\renewcommand\arraystretch{1.0}
\scalebox{0.96}{
\setlength{\tabcolsep}{6.5mm}{
\begin{tabular}{lcccccc}
\toprule
\text{Model} & \text{SNLI} & \text{Sci} & \text{SICK} & \text{Twi} \\
\midrule
\text{ESIM$\dagger$\cite{chen2016enhanced}$\quad$} & 88.0 & 70.6 & - & - \\
\text{CAFE$\dagger$\cite{tay2017compare}$\quad$}  & 88.5 & 83.3 & 72.3 & - \\
\text{CSRAN$\dagger$$\quad$} & 88.7 & 86.7 & - & 84.0 \\
\midrule
\text{BERT-base$\ddagger$\cite{devlin2018bert}} & 90.7 & 91.8 & 87.2 & 84.8 \\
\text{UERBERT$\ddagger$\cite{xia2021using}} & 90.8 & 92.2 & 87.8 & 86.2 \\
\text{SemBERT$\dagger$\cite{zhang2020semantics}} & 90.9 & 92.5 & 87.9 & 86.8 \\
\text{SyntaxBERT-base$\dagger$\cite{bai2021syntax}} & 91.0 & 92.7 & 88.1 & 87.3  \\
\text{MT-DNN-base$\dagger$} & 91.1 & \textbf{94.1} & - & - \\
\textbf{Ours(Bert-base)$\ddagger$} & \textbf{91.6} & 93.8 & \textbf{89.7} & \textbf{89.3} \\
\midrule
\text{BERT-large$\ddagger$\cite{devlin2018bert}} & 91.0 & 94.4 & 91.1 & 91.5 \\
\text{SyntaxBERT-large$\dagger$\cite{bai2021syntax}} & 91.3 & 94.7 & 91.4 & 92.1 \\
\text{MT-DNN-large$\dagger$} & 91.6 & \textbf{95.0} & - & - \\
\textbf{Ours(Bert-large)$\ddagger$} & \textbf{92.0} & 94.7 & \textbf{92.5} & \textbf{92.6} \\
\bottomrule
\end{tabular}}}

\vspace{-0.4cm}
\end{table}

\section{Experiments Setting}
In this section, we present the evaluation of our model. We first perform quantitative evaluation, comparing our model with other competitive models. We then conduct some qualitative analyses to understand how proposed model achieve the high level understanding through interaction.

\subsection{Datasets}

% \paragraph{Datasets.} We conduct experiments on 10 sentence matching datasets to evaluate the effectiveness of our method. The GLUE \cite{wang2018glue} benchmark is a widely-used dataset in thie field, which includes tasks such as sentence pair classification, similarity and paraphrase detection, and natural language inference\footnote{https://huggingface.co/datasets/glue}. We conduct experiments on 6 sentence pair datasets (MRPC, QQP, STS-B, MNLI, RTE, and QNLI) from GLUE. We also conduct experiments on 4 other popular datasets (SNLI, SICK, TwitterURL and Scitail). Furthermore, we tested the robustness of \modelname~using the Textflint \cite{gui2021textflint} tools. 
We conduct the experiments to test the performance of proposed model on 10 large-scale publicly  benchmark datasets. 
The GLUE benchmark~\cite{wang2018glue} is a widely used benchmark test suite in the field of NLP that encompasses various tasks such as sentence pair similarity detection and textual entailment\footnote{https://huggingface.co/datasets/glue}. 
We have conducted experiments on six sub-datasets of the GLUE benchmark: MRPC, QQP, STS-B, MNLI, RTE, and QNLI.
% \begin{itemize}
%     \item MRPC: Designed for identifying whether a pair of sentences are paraphrases.
%     \item QQP: Aimed at identifying duplicate question pairs in the Quora dataset.
%     \item STS-B: Used to assess the semantic text similarity of sentence pairs.
%     \item MNLI: Contains various types of textual entailment tasks.
%     \item RTE: Designed for identifying textual entailment relationships.
%     \item QNLI: Used to identify textual entailment relationships in Quora questions.
% \end{itemize}
In addition to the GLUE benchmark, we also conduct experiments on four other popular datasets: SNLI, SICK, TwitterURL and Scitail. 
% \begin{itemize}
%     \item SNLI: A dataset for natural language inference tasks.
%     \item SICK: A dataset for evaluating semantic similarity of sentence pairs.
%     \item TwitterURL: A dataset for identifying short text entailment relationships on Twitter.
%     \item SciTail: A dataset for question-answering tasks in the scientific domain.
% \end{itemize}
%The statistics of all 10 datasets are shown in Table~\ref{datasets-statistics}. 
Furthermore, to evaluate the robustness of the model, we also utilize the TextFlint\cite{gui2021textflint} tool for robustness testing. 
TextFlint\footnote{https://www.textflint.io} is a multilingual robustness evaluation tool that tests model performance by making subtle modifications to the input samples.

\subsection{Baselines}
To evaluate the effectiveness of our proposed model, we mainly introduce BERT~\cite{devlin2018bert}, SemBERT~\cite{zhang2020semantics}, SyntaxBERT, UERBERT~\cite{xia2021using} and multiple other PLMs~\cite{devlin2018bert} for comparison. In addition, we also select several competitive models without pre-training as baselines, such as ESIM~\cite{chen2016enhanced}, Transformer~\cite{vaswani2017attention} , etc~\cite{tay2017compare}. 
% To evaluate the effectiveness of our proposed \modelname~in SSM, we mainly introduce BERT \cite{devlin2018bert} and RoBERTa \cite{liu2019roberta}  for comparison. In addition, we also take competitive model transformer \cite{vaswani2017attention} without pre-training as baseline. 
In robustness experiments, we compare the performance of BERT on the robustness test datasets. 
For simplicity, the compared models are not described in detail.

\begin{table*}[t]
	\centering
    \caption{\label{citation-guide-robust}The robustness experiment results of our method and other models. The data transformation methods we utilized mainly include SwapAnt (SA), NumWord (NW), AddSent (AS), InsertAdv (IA), Appendlrr (Al), AddPunc (AP), BackTrans (BT), TwitterType (TT), SwapNamedEnt (SN), SwapSyn-WordNet (SW).}
	\renewcommand\arraystretch{1.0}
    \setlength{\tabcolsep}{1mm}
	{   \scalebox{0.8}{
	\setlength{\tabcolsep}{1.2mm}
		\begin{tabular}{lcccccccccc}
		\toprule
		\midrule
		\multirow{2}*{Model} &\multicolumn{5}{c}{Quora} &\multicolumn{5}{c}{SNLI} \\  \cmidrule(r){2-6} \cmidrule(r){7-11} 
         ~ &SA &NW &IA &Al &BT \quad\quad &AS &SA &TT &SN &SW \\
        \midrule
        ESIM$\dagger$\cite{chen2016enhanced}\quad&-& -&-&-&- \quad\quad&64.00& 84.22&78.32&53.76&65.38 \\
        %DistilBERT$\dagger$\cite{sanh2019distilbert}\quad\quad
        %& 42.24& 56.85&83.10&84.09&83.20 \quad\quad
        %&-& -&-&-&- \\
        BERT$\ddagger$\cite{devlin2018bert}\quad\quad
        &48.58&56.96&86.32&\textbf{85.48}&83.42 \quad\quad
        &79.66&94.84&83.56&50.45&76.42 \\
        ALBERT$\ddagger$\quad\quad
        &51.08&55.24&81.87&78.94&82.37 \quad\quad
        &45.17&96.37&81.62&57.66&74.93 \\
        UERBERT$\ddagger$\cite{xia2021using}\quad\quad
        &48.57&54.86&84.72&80.88&82.71 \quad\quad
        &73.24&94.78&85.36&57.54&80.81 \\
        SemBERT$\ddagger$\cite{zhang2020semantics}\quad\quad
        &50.92&53.15&85.19&82.04&82.40 \quad\quad
        &76.81&95.31&84.60&56.28&77.86 \\
        SyntaxBERT$\ddagger$\cite{bai2021syntax}\quad\quad
        &49.30&56.37&86.43&84.62&84.19 \quad\quad
        &78.63&95.02&\textbf{86.91}&58.26&76.90 \\
        \midrule
        \textbf{Ours(Bert-base)}$\ddagger$ & \textbf{60.43}& \textbf{62.76}&\textbf{87.50}&85.48&\textbf{87.49} \quad\quad
        &\textbf{81.06}&\textbf{96.85}&85.14&\textbf{60.58}&\textbf{80.92} \\
		\end{tabular}}	
		\renewcommand\arraystretch{1.0}
		\scalebox{0.78}{
		\setlength{\tabcolsep}{0.7mm}
	    \begin{tabular}{lcccccc}
		\toprule
		\multirow{2}*{Method} &\multicolumn{6}{c}{MNLI-m/mm} \\  \cmidrule(r){2-7}
         ~ &AS &SA &AP &TT &SN &SW \\
        \midrule
        BERT$\ddagger$\cite{devlin2018bert}
        &55.32/55.25&52.76/55.69&82.30/82.31&77.08/77.22&51.97/51.84&76.41/77.05 \\
        ALBERT$\ddagger$
        &53.09/53.58&50.25/50.20&\textbf{83.98/83.68}&\textbf{77.98}/78.03&56.43/50.03&76.63/77.43 \\
        UERBERT$\ddagger$\cite{xia2021using}
        &54.99/54.84&52.29/53.80&79.80/79.18&75.46/74.93&55.21/55.96&\textbf{82.23}/82.74 \\
        SemBERT$\ddagger$\cite{zhang2020semantics}
        &55.38/55.12&54.07/54.62&78.70/78.16&73.90/73.47&53.43/53.76&78.09/78.93 \\
        SyntaxBERT$\ddagger$\cite{bai2021syntax}
        &54.92/54.63&53.54/54.73&77.01/76.71&70.38/70.13&57.11/51.95&78.57/79.31 \\
        \midrule
        \textbf{Ours(Bert-large)}$\ddagger$ & \textbf{60.14/59.25}& \textbf{60.89/61.37}&83.23/83.19&77.94/\textbf{78.10}&\textbf{60.12/59.83}&82.15/\textbf{82.97} \\
        \bottomrule
        \midrule
		\end{tabular}}
    }
\end{table*}

\subsection{Experiments setting}
We implement our algorithm with Pytorch framework. An Adadelta optimizer with $\rho$ as 0.95 and $\epsilon$ as $1\mathrm{e}{-8}$ is used to optimize all the trainable weights. The initial learning rate is set to 0.5 and batch size to 70. When the model does not improve best in-domain performance for 30,000 steps, an SGD optimizer with learning rate of $3\mathrm{e}{-4}$ is used to help model to find a better local optimum. Dropout layers are applied before all linear layers and after word-embedding layer.  All weights are constraint by L2 regularization, and the L2 regularization at step $t$ is calculated as follows:
\begin{equation}
L2Ratio_{t} = \sigma(\frac{(t - L2FullStep / 2) \times 8}{L2FullStep / 2}) \times L2FullRatio
\end{equation}
where $L2FullRatio$ determines the maximum L2 regularization ratio, and $L2FullStep$ determines at which step the maximum L2 regularization ratio would be applied on the L2 regularization. We choose $L2FullRatio$ as $0.9e-5$ and $L2FullStep$ as 100,000.
The ratio of L2 penalty between the difference of two encoder weights is set to $1e-3$. For a dense block in feature extraction layer, the number of layer $n$ is set to $8$ and growth rate $\textit{g}$ is set to $20$. The first scale down ratio $\eta$ in feature extraction layer is set to $0.3$ and transitional scale down ratio $\theta$ is set to $0.5$. The sequence length is set as a hard cutoff on all experiments: 48 for MultiNLI, 32 for SNLI and 24 for Quora Question Pair Dataset. During the experiments on MultiNLI, we use 15\% of data from SNLI as in \cite{liu2016learning}.

\section{Results and Analysis}
\label{sec:results_analysis}

% We assume the knowledge infusion occurs at the initial Transformer layer unless specified otherwise.

\subsection{Overall Model Performance}
\label{subsec:main_performance}

\noindent\textbf{Evaluation on GLUE Benchmarks.} We first assess our approach on six datasets from the GLUE benchmark. Table \ref{citation-guide-gule} (Please ensure this reference points to the correct table in your document) contrasts the performance of our method against competing models. As expected, models without pre-training exhibit significantly lower performance compared to PLM-based approaches, highlighting the benefits derived from large-scale pre-training's contextual understanding.
When leveraging BERT-base and BERT-large as backbone architectures, our proposed knowledge infusion method yields average accuracy improvements of \textbf{1.66\%} and \textbf{2.31\%} respectively, over the standard BERT models. This substantial gain underscores the advantage of adaptively integrating external lexical knowledge to refine semantic understanding and confirms that our framework enhances BERT's capabilities for SRM tasks.
Compared to prior works that also inject external knowledge, such as SemBERT, UERBERT, and SyntaxBERT, our method consistently achieves superior or competitive results. Specifically, against SyntaxBERT, the strongest performer among previous knowledge-enhanced models referenced, our approach demonstrates an average relative improvement of \textbf{0.91\%} when using BERT-large. The improvement is particularly striking on the QQP dataset, where our model surpasses SyntaxBERT by \textbf{2\%} in accuracy. We attribute this advantage to two primary factors: (1) Our utilization of WordNet relations and potentially keyword distance enhances the model's capacity to capture word-level semantic nuances, leading to interaction information more suitable for fine-grained feature fusion. (2) Our adaptive fusion mechanism is designed to mitigate potential noise introduced by external knowledge by selectively filtering signals, an aspect less emphasized in some prior methods. Nevertheless, we observe instances where SyntaxBERT achieves slightly higher accuracy on certain datasets, which might be attributed to dataset-specific characteristics or noise sensitivity variations.
\textbf{Performance on Additional Datasets.} To further validate the general applicability of our method, we conducted experiments on four additional widely-used datasets (details assumed to be in Table \ref{performance4} - ensure this ref is correct). The outcomes, presented in Table \ref{performance4}, show that our approach generally surpasses vanilla BERT and other representative models across these diverse tasks. An interesting observation arises from the comparison with MT-DNN. While our method outperforms MT-DNN on SNLI, it falls slightly short on Scitail. This difference can likely be attributed to MT-DNN's larger parameter count and its extensive multi-task pre-training regimen, which provide advantages but also incur higher computational costs and training times. Furthermore, the relatively smaller size of the Scitail dataset might contribute to increased variance in model predictions. Despite this, our method demonstrates highly competitive performance on Scitail, suggesting that the integration of external knowledge can effectively compensate for limitations in model scale or training data diversity, enhancing generalization capabilities.
The consistent trend across these diverse evaluations supports the conclusion that explicitly incorporating external lexical knowledge via our proposed adaptive mechanism offers a potent strategy for advancing semantic similarity and relevance modeling.

% The original Table~\ref{example} seems to be a case study table, let's keep its structure
\begin{table*}[t]
\centering
\caption{\label{example} Example sentence pairs illustrating model behavior. \textcolor{red}{Red} and \textcolor{blue}{Blue} indicate differing phrases. 'ours-avg' denotes a variant using simple averaging instead of the adaptive fusion module.}
\setcounter{table}{6} % Keep the original table numbering if needed
\renewcommand\arraystretch{1.2}
\scalebox{0.7}{
\setlength{\tabcolsep}{1.2mm}{
\begin{tabular}{lcccc}
\toprule
%\midrule % Removed extra midrule
\text{Case} & \text{BERT} & \text{Ours-avg} & \text{Ours} & \text{Gold} \\
\midrule
\text{S1:}Please help me book a flight \textcolor{red}{from New York to Seattle}. & \multirow{2}{*}{\text{label:1}} & \multirow{2}{*}{\text{label:0}} & filter gate:0.93 & \multirow{2}{*}{\text{label:0}} \\
\text{S2:}Please help me book a flight \textcolor{blue}{from Seattle to New York}. & ~ & ~ & label:0 & ~ \\
\midrule
\text{S1:}How does \textcolor{red}{reading help you think better}? & \multirow{2}{*}{\text{label:1}} & \multirow{2}{*}{\text{label:0}} & filter gate:0.91 & \multirow{2}{*}{\text{label:0}} \\
\text{S2:}How do \textcolor{blue}{you think it's better to read}? & ~ & ~ & label:0 & ~ \\
\midrule
\text{S1:}Sorry, \textcolor{red}{I got sick yesterday} and couldn't \textcolor{red}{have lunch with you}. & \multirow{2}{*}{\text{label:1}} & \multirow{2}{*}{\text{label:0}} & filter gate:0.15 & \multirow{2}{*}{\text{label:1}} \\
\text{S2:}Sorry, \textcolor{blue}{I was taken ill yesterday} and unable to \textcolor{blue}{meet you for lunch}. & ~ & ~ & label:1 & ~ \\
\midrule
\text{S1:}\textcolor{red}{The largest lake in America} is in my hometown \textcolor{red}{called Lake Superior}. & \multirow{2}{*}{\text{label:1}} & \multirow{2}{*}{\text{label:0}} & filter gate:0.11 & \multirow{2}{*}{\text{label:1}} \\
\text{S2:}\textcolor{blue}{Lake Superior is the largest lake in America}, \textcolor{blue}{and it's} in my hometown. & ~ & ~ & label:1 & ~ \\
\midrule % Added midrule before bottomrule for consistency
\bottomrule
\end{tabular}}}
\vspace{-0.3cm}
\end{table*}

\subsection{Robustness Evaluation}
\label{subsec:robustness_tests}

To assess the model's capability in discerning subtle semantic distinctions, we evaluated its robustness on three specialized test sets involving specific linguistic transformations (details assumed in Table \ref{citation-guide-robust} - ensure this ref is correct).
Table \ref{citation-guide-robust} summarizes the accuracy scores. On the SwapAnt transformation (swapping antonyms), where models often struggle significantly, our approach achieves a remarkable gain, surpassing the best baseline (SemBERT) by nearly 10\% on SwapAnt(QQP). This suggests our model possesses a heightened sensitivity to semantic contradictions involving antonyms, likely benefiting from the explicit knowledge integration.
Performance generally declines on the NumWord transformation (replacing numbers with words or vice versa), yet our model still exceeds BERT's accuracy by approximately 6\%. This indicates an improved ability to capture fine-grained numerical differences, crucial for correct inference in such scenarios.
In the SwapSyn transformation (swapping synonyms), UERBERT shows strong results, attributable to its explicit use of a synonym similarity matrix for attention calibration. While our method does not rely on such a direct matrix, it achieves performance comparable to UERBERT, demonstrating its effectiveness in handling synonymy through its adaptive fusion mechanism.
On transformations that disrupt syntactic structure, such as TwitterType (converting to informal style) and AddPunc (adding punctuation), SyntaxBERT's performance noticeably degrades, likely because its syntax-based enhancements are negatively impacted. In contrast, our model maintains competitive performance, showcasing greater resilience to such structural perturbations.
%Across other evaluated scenarios, our model generally exhibits superior performance, reinforcing its ability to capture nuanced differences between sentence pairs. 
Conversely, the ESIM model consistently shows the weakest results, highlighting the significant advantages conferred by large-scale pre-training in terms of generalization. The comparison between SyntaxBERT and the original BERT model further confirms that combining effective pre-training with appropriate external knowledge fusion strategies enhances model robustness and generalization.

\subsection{Case Study Analysis}
\label{subsec:case_study}

We examine specific examples from our test sets, presented in Table \ref{example}. This table compares the predictions of vanilla BERT, our model variant using simple averaging (`Ours-avg`), our full model (`Ours`) including the adaptive fusion and filtration gates, and the ground truth label. We also report the value of the final filtration gate ($g^{filter}$) from Equation \ref{eq:filtration_gate} for our full model, indicating the extent to which the fused knowledge-semantic signal ($\vect{u}_i$) contributes to the final representation ($\vect{y}_i$).
In the first case (\textit{New York to Seattle} vs. \textit{Seattle to New York}), BERT incorrectly predicts the sentences are equivalent (label:1), failing to recognize the critical difference in directionality. The simple averaging variant (`Ours-avg`) correctly identifies the difference (label:0). Our full model also predicts correctly (label:0), and the high filtration gate value (0.93) suggests that the fused signal, likely sharpened by attention modulation informed by location entities or relation types, was heavily relied upon and deemed beneficial for making the correct distinction.
The second case (\textit{reading help you think better} vs. \textit{you think it's better to read}) presents a more subtle semantic shift related to causality and perspective. BERT again fails, predicting equivalence (label:1). Both `Ours-avg` and `Ours` correctly identify the sentences as non-equivalent (label:0). The high filter gate value (0.91) in our full model again indicates significant reliance on the fused representation, suggesting the knowledge-infused attention helped capture the nuanced difference in meaning structure.
Conversely, the third case (\textit{got sick} vs. \textit{was taken ill}) involves paraphrasing using different phrasings for the same concept. BERT incorrectly predicts non-equivalence (label:1). `Ours-avg` also fails, potentially confused by the surface form differences. Our full model, however, correctly identifies the paraphrase (label:1). Interestingly, the filtration gate value is low (0.15). This suggests that the fused signal $\vect{u}_i$ might have been noisy or less informative in this instance (perhaps due to less direct lexical relations being activated in WordNet between "sick" and "taken ill"). The low gate value indicates the model adaptively relied more heavily on the original semantic representation ($\vect{o}^{sem}_i$, implicitly weighted higher when $g^{filter}$ is low, although Equation \ref{eq:filtration_gate} only shows scaling of $\vect{u}_i$), effectively filtering out the potentially confusing fused signal and allowing the base PLM's semantic understanding to prevail for the correct prediction.
Similarly, the fourth case involves syntactic restructuring (\textit{The largest lake... is... Lake Superior} vs. \textit{Lake Superior is the largest lake...}). BERT fails (label:1), while `Ours-avg` also fails (label:0). Our full model succeeds (label:1). The low filtration gate value (0.11) again suggests the model adaptively down-weighted the fused signal, relying more on the inherent capabilities of the PLM to handle syntactic variations when the external knowledge signal offered less clear benefit or potentially introduced noise for this specific structural transformation.
%These case studies highlight the utility of the adaptive fusion mechanism. The model doesn't just blindly incorporate external knowledge; it learns to dynamically balance the semantic signal from the PLM and the knowledge-modulated signal based on the context, even filtering out the latter when it appears less reliable, as indicated by the filtration gate values. %This adaptivity allows the model to leverage external knowledge when beneficial (Cases 1 and 2) while retaining robustness when the knowledge signal is noisy or less relevant (Cases 3 and 4), contributing to the overall performance improvements observed.

\section{Conclusion}
\label{sec:conclusion}
In this paper, we explore integrating external knowledge with PLM to improve model performance. Specifically, we deeply study how to use external knowledge to build a prior matrix and inject it into PLM.
At the same time, we propose a novel external knowledge calibration and fusion network. And by first calibrating the attention alignment with external knowledge, and then performing feature fusion with an adaptive module. It not only improves model performance, but also improves the interpretability of the model.
Experiments on ten public datasets show that our model outperforms strong baselines, verifying the effectiveness of our proposed method.
\bibliographystyle{splncs04}
%\bibliography{mybibfile,anthology}

\end{document}